# Context-dependent Explainability and Contestability for Trustworthy Medical Artificial Intelligence: Misclassification Identification of Morbidity Recognition Models in Preterm Infants

Işıl Güzey[a], Özlem Uçar[a], Nükhet Aladağ Çiftdemir[b], Betül Acunaş[b]

[a]Trakya University Faculty Engineering, Department of Computer Engineering, Edirne-Turkey

[b]Trakya University Faculty of Medicine, Division of Neonatology, Pediatric Department, Edirne-Turkey

**Abstract**

Although machine learning (ML) models of AI achieve high performances in medicine, they are not free of errors. Empowering clinicians to identify incorrect model recommendations is crucial for engendering trust in medical AI. "Explainable AI (XAI)" aims to address this requirement by clarifying AI reasoning to support the end users. Several studies on biomedical imaging achieved promising results recently. Nevertheless, solutions for models using tabular data are not sufficient to meet the requirements of clinicians yet. This paper proposes a methodology to support clinicians in identifying failures of ML models trained with tabular data. We built our methodology on three main pillars: *decomposing the feature set by leveraging clinical context latent space, assessing the clinical association of global explanations, and Latent Space Similarity (LSS) based local explanations.* We demonstrated our methodology on ML-based recognition of preterm infant morbidities caused by infection. The risk of mortality, lifelong disability, and antibiotic resistance due to model failures was an open research question in this domain. We achieved to identify misclassification cases of two models with our approach. By contextualizing local explanations, our solution provides clinicians with actionable insights to support their autonomy for informed final decisions.

Static features (e.g., age, gender, and genetic variations) are often used in medical ML models with dynamic (e.g., blood test results and vital signs) features. Although there are cases of latent patterns in physiology based on relations of static and dynamic features to distinguish health and disease states, ML models do not consider this distinction in the training stage. Therefore, we hypothesized that challenging a model decision by leveraging LSS-based similarity to ensure the existence of an expected latent pattern could facilitate identifying when models fail. Our methodology can be translated to other medical domains with consideration of the clinical context. Additionally, it has the potential to facilitate the model audit and improvement endeavors based on feedback from clinicians.

# 1. Introduction

A remarkable number of research studies have demonstrated that machine learning (ML) algorithms of AI may achieve promising clinical decision support performance by automatically detecting patterns in healthcare data [1–4]. However, these high-performing ML algorithms are generally opaque, like black boxes in their inner processes. On the other hand, clinicians need to understand how an algorithm reaches a specific decision to translate it into clinical practice and conclude that it is trustworthy. This translation is also considered challenging in clinical settings for transparent algorithms like logistic regression and decision trees due to the computational complexity of high-dimensional health data.

ML algorithms' capability of detecting patterns in high-dimensional data more accurately and speedily than humans comes at a cost. To achieve the best possible performance, they use not only the intended domain features but all the information within the dataset, such as spurious correlations and inherent bias, as in [5], which compromise their trustworthiness.

There are also several impressive examples of this phenomenon from the healthcare domain. An ML model to detect high-risk pneumonia patients classified patients with asthma as low-risk [6]. The asthmatic patients in the used dataset with pneumonia were admitted directly to the ICU (Intensive Care Unit). Thanks to the aggressive care they received, their risk of death due to pneumonia was lower than the general population. Another study detected that surgical skin markings interfered with the deep learning model's diagnosis capability by increasing the melanoma probability score and causing false-positive rates [7].

To address this understandability and trust issues, plenty of research studies focused on the explainability of models in recent years for all the domains using ML. Diverse conceptual and methodological approaches to explain the decision logic of the models have been developed and discussed under the Explainable Artificial Intelligence (XAI) topic [8,9].

The 'model-agnostic' or model-independent methods developed within the scope of XAI, which apply to any model, have been chiefly preferred as they enable the comparison between models[10]. While global explanations of these approaches contributed to the overall model understanding, the trust issue was not sufficiently addressed for instance-based (local explanations), as they generally do not practically clarify the causal relationship of the real world but explain the inner workings of models. Besides, several studies state that recognizing algorithmic advice errors and improving the models are crucial aspects of trustworthy artificial intelligence in medicine [11–14].

For high-stakes decision domains, including medicine, contestability was raised as an indispensable requirement [15–18] to ensure the principle of non-maleficence and adherence to legal requirements of the European Union's General Data Processing Regulation (GDPR) in Article 22(3), which states that data subjects have the right to contest decisions based solely on automated processing [19].

Algorithmic research studies for early recognition of late-onset sepsis (LOS) and necrotizing enterocolitis (NEC) in preterm infants is a topic that is in great need of a contestability concept. Early recognition and immediate antibiotic-based treatments are the only measures to reduce neonatal morbidity and mortality caused by these diseases. Nevertheless, these are challenging tasks for clinicians due to subtle clinical signs before the abrupt deterioration of patients. Although various algorithmic models, based on data derived from monitored non-invasive vital signs, achieved good recognition performances, confusion due to overtrust in these models has serious risks. While trusting a false-negative model decision causes a delay in antibiotic therapy, trusting a false-positive decision has the risk of unnecessary antibiotic exposure in the noninfected infant, which may contribute to the development of antibiotic resistance [20].

To address this problem, we developed a methodology for clinicians to assess ML models' decision logic and practically identify misclassifications. The steps of our methodology are as follows:

- The overall decision logic must be consistent with the domain knowledge as the first step to ensure the reliability of a model. If it strongly correlates with an irrelevant feature, such as the snow in the background, as in [5], we can infer a severe problem with the training process. Withdrawal of the model may be the most reasonable choice in such a case. On the other hand, if model predictions appropriately rely on the features, which are also clinically meaningful in the context, the instance-based model decisions may be evaluated as the next step.

- For instance-based model reliability, we base our approach on misclassification inspection by checking whether the decision stands on reasonable facts from the perspective of clinicians. We leverage the domain knowledge, the common variations of dynamic features for healthy and ill infants, concerning their similarities within latent space constituted by maturity and gender static features.

- We present both steps of our analysis with parsimonious visualizations, abstaining from numerical performance results, as their validity highly depends on data distribution for a specific dataset. Thus, with interactive facilities as access to the clinical details and trajectories of similar patients, clinicians may decide with autonomy utilizing the distilled information instead.

To our knowledge, there is no study to identify the model misclassifications of either preterm morbidity recognition models or any other clinical ML model using tabular data, despite the widespread availability of Electronic Health Records (EHRs) [21]. In addition to proposing a methodology for these shortcomings as our contributions, we aim to highlight the importance of collaborative involvement of engineering and medical perspectives for trustworthy medical AI.

## 2. Related Work
### 2.1 Explainability, Contestability, and Trust in Medical Domain

The global explanations of model-agnostic XAI methods help to understand a model's overall functioning and auditing domain context coherence, such as clinical validation in the medical domain. Nevertheless, the informative potential of the model-agnostic local explanations has recently become a serious debate for high-stake decisions. These methods generally focus on explaining the decision logic of an instance with an approach like interpreting regional mathematical associations of feature values and class labels [5,22,23]. As nicely analogized in [24], "Machine learning is itself a correlation box" and this approach neither ensures algorithmic decisions' legitimacy nor provides a practical means for justifiability and contestability needed for high-stake decisions [15].

Indeed, a survey [25] to investigate requirements from explainability to improve clinicians' trust in ML models revealed that they viewed explainability as a means of justifying their clinical decision-making. Carefully designed information visualization, parsimonious and actionable steps were also expressed as required facilities to supplement clinical models. The model performance or prediction certainty scores were also perceived as complicated by clinicians in this survey.

In the literature, we came across several studies focusing on the need for model decision justification and information visualization as a means of explainability. In [26], a tool is designed for physicians to understand AI-based analysis of chest X-ray images. Depending on physicians' need for more information, as informative comparisons, instead of vague descriptions of abnormalities, they visually presented contextualized observations with contrastive examples and comparisons of images across patients having the same observation to "justify how two images are clinically (dis)similar". Although the explained system was not a black-box model, a case-based reasoning system (CBR) [27] presents visual explanations, which display the dimension names and their associated values to explain why similar cases are similar to the query case and on which dimensions, to justify the results of breast cancer classifications. [28] complements the neural network with a CBR that can find explanatory cases for an explanation-by-example justification of model predictions and presents the values of the relevant (important) features between the given query and the three most similar cases.

Considering the transparency needs of models in radiology, [29] proposes a pervasively appropriate paradigm of proximity for opening black-box models by presenting annotated cases proximal to the instance to justify its model decision as an explanation. Furthermore, they emphasized the utility of this approach to detect the cases poorly addressed by the model; thus, giving developers feedback would contribute to model improvement endeavors.

The common aspect of these studies is resorting to similarity-based reasoning concepts to justify a model decision, which is considered within the scope of the explanation-by-example or prototype-based XAI approach [8,9]. As a matter of fact, due to concerns of cognitive overload risk and impracticality, the similarity-based model evaluation approach has not been considered appropriate in medicine, particularly for ICU settings [25]. On the other hand, these studies [26–28] cope with this shortcoming by leveraging contextualization, principal component analysis (PCA) and multidimensional scaling (MDS), and feature importance, respectively.

The latent space concept is another approach to cope with this dimension complexity of similarity-based evaluation [30]. As examples from medicine, [31] relates meaningful latent concepts to prediction targets and observes data for domain-specific supporting evidence, and [32] focuses on identifying the misclassification of chest x-ray images considering patient subgroups and misclassification predictors. Medical imaging is the main domain where latent space exploration was considered for the explainability of deep learning models exploiting saliency maps and encoder–decoder-based architectures [33–35].

In contrast with the widespread availability of Electronic Health Records (EHRs) for clinical settings, explainability studies for tabular data are generally limited to investigating the global relationships between classification and features. Besides, there is a lack of studies addressing the needs of clinicians in context-dependent explanations and investigating model failures to support informed decision-making while keeping their reasoning interfaces similar to their mental models.

Essentially, static features such as age, gender, and genetic variations [36–41] affect the normative values of physiological (dynamic) features independent from health and disease states. For the case of preterm infants, there is a significant pattern of dynamic feature-based variation for healthy and ill preterm infants concerning maturity and gender-based static features latent space, as we will explain in section 2.2 in more detail. Therefore, we hypothesized misclassifications might be identified due to (dis)similarities and support clinicians in justifying or contesting the patient-based decisions of the ML models.

While contestation and justification concepts are used depending on the explanation requirement of a specific decision, the contestability term is used as the capability of a system to challenge algorithmic decisions [24], and explainability is considered an enabler for contestability [16,42].

**2.2 LOS and NEC in preterm infants**

Late-onset neonatal sepsis (LOS), which occurs after the third day of life, is a multi-organ complication triggered by an immunological response to the infection [43]. Specific clinical signs appear only late when the severe illness is already present [44]. Blood cultures are sent for testing once the signs and symptoms develop and treatment starts. Nevertheless, laboratory results can take up to a day or more and are associated with a significant number of false negatives and positives.

Based on the hypothesis that "abnormal HRC (Heart Rate Characteristics) might precede the clinical diagnosis of neonatal sepsis," the results of early research have confirmed that even hours to a day before the symptoms appear, decreased variability and decelerations of heart rate (HR) were observed in infants with LOS [45]. Onwards this study, in a randomized controlled clinical trial of 3003 very low birth weight infants with post-menstrual age < 33 weeks, by continuously monitoring an index, HeRO score, based on heart rate variability (HRV) metrics, the mortality rate was reduced from 10.2% to 8.1% [46].

The underlying fact of this phenomenon has roots in the immature autonomic nervous system (ANS) of preterm infants, which optimally needs 37 weeks of intrauterine development [47]. The HRV, the beat-to-beat variation in either heart rate or the duration of the R–R interval–the heart period, is a measure in assessing the ANS. Its increase is linked to good health and decreased stress levels [48]. Several studies showed that preterm infants without a specific morbidity condition had decreased HRV compared to healthy full-term infants [49]. However, it increased in correlation with infants' gestational age, weight, and post-natal age [50,51]. Nevertheless, the suboptimal maturation of ANS negatively affects the capability of adapting to a continually changing environment and makes the preterm infant more vulnerable to pathological conditions like infection, with an increased risk of severe morbidity and mortality [52].

Besides ANS maturity, the male gender has been noticed to be another factor for preterm infants regarding adverse disease outcomes. Therefore, it is highly recommended when designing clinical and experimental research [53].

In addition to HRV and demographic parameters, several other vital sign parameters and their derivations have also been used in algorithmic research studies of the domain [54]. In [55,56], the cross-correlation of HR and oxygen saturation (SpO2) was found to be the best illness predictor out of the metrics derived from respiratory rate, HR, and SpO2. Recognition of NEC, which resembles mostly the clinical picture of LOS, was also included within the scope of these studies. We also have a few preliminary studies for this research topic [57,58]. Recently, respiratory characteristics, infant activity [59,60], and visibility graph features derived from the heart rate time series, in addition to HRV-based metrics [61], were used in ML-based LOS recognition studies. [62] used the HeRO score and clinical parameters connected to the risk of neonatal sepsis in their ANN model that outperformed the diagnostic performance of only the HeRO score, with a reduction of false positives on their dataset. To our knowledge, there is only one XAI study for neonatal sepsis [63], but there is no contestability-based study within the domain.

## 3. Materials & Methods
### 3.1 Patients and Clinical Data Collection

We conducted our study with the data of patients hospitalized in the tertiary care (Neonatal Intensive Care Unit) NICU of Trakya University Hospital, Turkey, from February 2017 to October 2017. We targeted the study group as preterm infants with Ga ≤ 32 weeks. Congenital abnormalities, heart disease, and inotropic support were considered exclusion criteria. With the local ethics committee approval (TUTF-BAEK2017\09), we collected already monitored HR and SpO2 pulse oximetry (PO) time-series data at two one-hour epochs daily (morning-11:00-12:00 and afternoon-16:00-17:00), when no care was given to infants.

Culture-positive LOS was diagnosed if positive blood culture results yielded a pathogen, and Clinical LOS if cultures were negative, but clinical signs and symptoms and acute phase reactants (C-reactive protein) implied clinical sepsis after three days of life [64]. We defined the diagnosis of NEC according to Bell's staging criteria modified by Walsh and Kliegman [65]. Los-Nec group consisted of episodes of 24 infants for the days when either LOS or NEC was diagnosed. We conducted a rigorous sample selection procedure for the control (healthy) group of 24 infants. For all the collected data, we reviewed the vital sign time series data and discarded episodes with artifacts. We targeted to ensure the maturity and gender-based balance of both groups. For each infant from the LOS-NEC group, we selected the artifact-free daily episode of an infant without any morbidity condition three days before and after, having similar gestational age, birth weight, weight, post-natal age, and gender for the control group. The demographic and clinical characteristics of the study population of 48 infants are presented in Table 1.

We collected vital signs from the Mindray iMec Patient Monitoring database of the NICU Central Monitoring System and transferred them to Mysql 8.0 Database on an Intel Core i5-8265U CPU @ 1.60GHz 1.80 GHz PC. We carried out the data exploration, cleaning, and feature extraction processes using Matlab R2019b version. We used R v 4.0.3 and R Studio v1.2.5019 for ML analysis on our dataset.

**Table 1.** Demographics of Study Population.

|  | Culture+ LOS | Clinical LOS | NEC | Healthy | Study Population |
|---|---|---|---|---|---|
| **Number of Infants** | 7 | 8 | 9 | 24 | 48 |
| **Ga (weeks)** | Median (28) Range (26, 31) | Median (29) Range (24, 30) | Median (29) Range (25, 32) | Median (28) Range (27, 31) | Median (28) Range (24, 32) |
| **Bw (gr)** | Median (915) Range (860, 1170) | Median (887) Range (535, 1325) | Median (1035) Range (845, 1565) | Median (1162) Range (690, 1570) | Median (1022) Range (535, 1570) |
| **W (gr)** | Median (1005) Range (770, 1310) | Median (1040) Range (525, 1780) | Median (1260) Range(865, 1730) | Median (1117) Range (685, 1800) | Median (1117) Range (525, 1800) |
| **Pna (weeks)** | Median (13) Range (8, 39) | Median (15) Range (4, 38) | Median (12) Range (5, 28) | Median (11) Range (5, 40) | Median (12.5) Range (4, 40) |
| **Female Male** | 4 3 | 3 5 | 3 6 | 15 9 | 25 23 |

## 3.2 Demographics and Vital Sign-based Features

We constituted our feature set with patient demographics and vital sign-based information (Table 2). We derived features from every 1 second (1 Hz) HR and SpO2 vital sign data, which were also considered in previous similar research within the context of algorithmic LOS and NEC recognition studies.

**Table 2.** List of all features with descriptions.

| Category | Feature | Description |
|---|---|---|
| **Demographics** | Gen | Gender |
| | Ga | Gestational Age |
| | Bw | Birth Weight |
| | W | Weight |
| | Pna | Post Natal Age |
| **Vital Signs** | XC-hr-spo2 | The max cross-correlation of HR and Spo2 |
| | SA-hr | Sample asymmetry of HR histograms |
| | Hrm | HR mean |
| | Spo2m | Spo2 mean |
| | Hs | % of Hypoxia duration |
| | Brs | % of Bradycardia duration |
| | Ts | % of Tachycardia duration |

Two features, the maximum cross-correlation between standardized HR and SpO2 and sample asymmetry of HR histogram, are indicators of ANS activation dysfunction, and their increase indicates morbidity risk. Cross-correlation between standardized HR *and Spo2* represents immature breathing patterns during illness, as changes in HR and SpO2 occur in synchrony with breathing pauses due to apnea associated with bradycardia and oxygen desaturation [56]. Sample asymmetry of HR histogram describes changes in the shape of the histogram of HR, which are caused by reduced accelerations and transient decelerations [66]. We calculated these feature values per the same methodologies in [56] and [66]. Mean HR and mean SPo2 are the arithmetic means of one-hour epochs. Hypoxia percentage is the percentage of seconds in which Spo2 is below 80%, and Bradycardia percentage and tachycardia percentage are the percentage of seconds in which HR value is below 85 and above 180 within the epoch, respectively. The daily episode values to be included in the dataset were taken as the average of both morning and afternoon 1-hour epochs.

### 3.3 Explainability and Contestability Approach

To understand model decision logic and identify model misclassifications, we focused on three research questions (RQs) determined by the clinicians in our team.

- *RQ 1: What is the overall decision logic of the models?*

The dataset-level feature importance analysis shows how much a model relies on each of the features [67], and the partial dependence plot (PDP) shows how the expected value of model prediction changes as a function of one or more features [68]. Both methods are commonly used to assess the overall decision logic of models [10,69]. While we considered all features in feature importance analysis as a general practice, we evaluated the PDP analysis with a domain-specific approach.

Several studies emphasize the importance and model performance contribution of infant maturity and gender static features [60,70]. This is an expected outcome in line with the clinical mechanisms, as explained in Section 2.2. The effect of static features comes into prominence along with the dynamic features representing the physiological state of an infant. Therefore, as a differentiation from general practice, we only focused on the dynamic ones in our one-dimensional PDP analysis instead of all the features for clarity. In two-dimensional PDP analysis, we targeted to explore a static feature along with a dynamic one to inspect their joint contribution to the model prediction.

- *RQ 2: Regarding the patient-specific decision of a model, what are dynamic feature values and ground truth classes for infants having similar maturity and gender?*

The main challenge of clinicians using an algorithmic recognition model in clinical practice is whether or not to trust its decision for a specific patient, even though the overall decision logic of the model is reliable. At this stage, one option was to analyze model-agnostic local instance-based explanations of the model to check that they are consistent with the domain knowledge. In response to a preliminary analysis of various model-agnostic local explanation methods, our team's clinician counterparts recommended using another approach instead of those methods. It was to assess the dynamic feature values of an infant concerning several other similar ones to be able to notice clues of error [12].

In high-risk contexts, reducing uncertainty by matching the situation with similar past experiences is a preferred decision-making method [71]. Clinicians frequently use this approach in their daily practices and favor similarity-based explanation classes but expect the similarity to be contextualized to their mental models [72]. In the neonatology domain, clinicians frequently use maturity (gestational age, post-natal age, and weight) and gender-based similarity assessment in the clinical context. There is already an HRV-based ANS maturation profiling study of healthy full-term infants from birth to 2 years considering physiologic differences between genders [73]. On the other hand, this profiling is complicated in preterms due to the cumulative individual effects of factors, including but not limited to gestational age, weight, and post-natal age. Poor ANS activity for this population is a sign of immaturity and morbidity.

Therefore, to dissociate immaturity and morbidity risk attributions, we selected an instance's maturity and gender-based similars from our dataset, along with their ground truth class values. Thus, we constitute a base for a ground truth explanation-by-example method [74], with a causal understanding associated with the context of the domain [75]. As gender is a qualitative one within this group of features, we used the Gower method [76] to find the neighbors. Eventually, we determined gestational age, post-natal age, weight, and gender as the sub-feature space dimensions of our latent space [77].

- *RQ 3: Can the results be seen over a <u>simple</u> visual interface?*

Particularly Intensive Care Unit (ICU) clinicians are exposed to massive amounts of information about patients, medical procedures, treatments, and other clinical data to be used in their systematic clinical decision processes. For clinicians to accept a new system in such a setting, it needs to be worthy of its cognitive load in novelty and actionability [78], reduces the quantity, and improve the consumability of information [79]. We applied this simplification by reducing four features; *gestational age*, *weight*, *post-natal age*, and *gender*, to just a single metric; the Gower distance. We leveraged this metric to identify (dis)similarities within Latent Space.

We prepared two-dimensional graphics for each of the two important dynamic features so that clinicians can analyze the variations of feature values and similarities of the instance and its neighbors with ground-truth labels. Figure 1 illustrates the framework of our explainability and contestability approach. The global explanations module corresponds to answering RQ1, and the local-explanations module to RQ 2 and RQ 3. This framework also depicts how our context-based explainability and contestability approach supports the model evaluation and improvement processes.

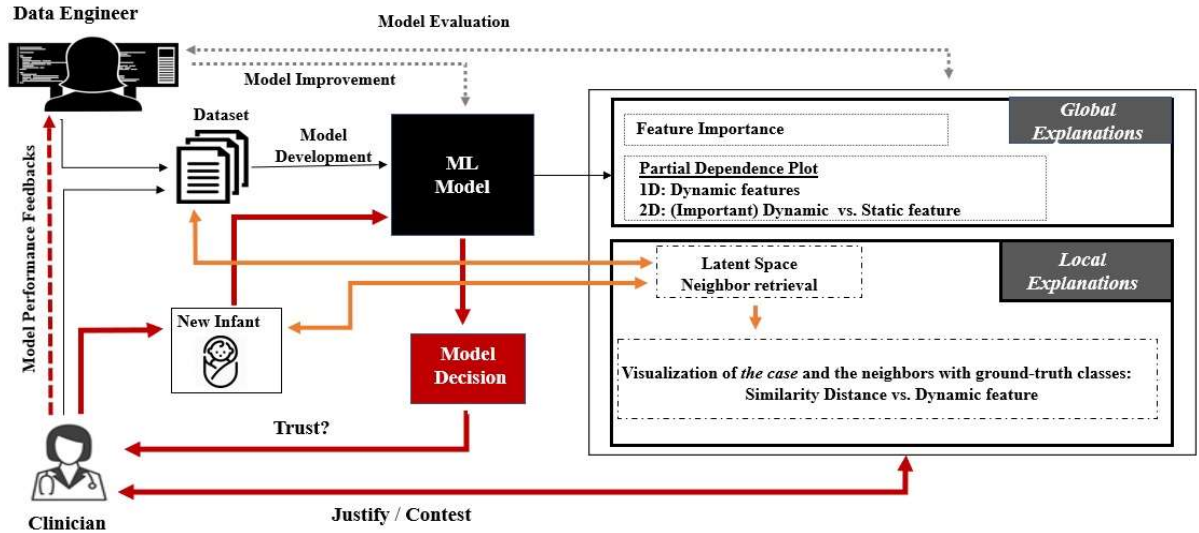

**Fig. 1**. The framework of the proposed explainability and contestability approach.

## 4 Results

### 4.1 Model Training and hyperparameters

As the feature importance of the models would be evaluated within the scope of the global explanations, and considering that it may be affected by multicollinearity [80], we iteratively removed highly correlated features from the dataset by evaluating the variance inflation factor (VIF) [81] of our dataset using the R statistical software, package "car" [82]. To ensure recommended maximum VIF threshold of 2.5 [83], we removed birth weight, tachycardia percentage, and hypoxia percentage features. We used R packages caret [84], DALEX[85], iml [86], and gower [87] for ML training, feature importance, PDP, and neighbor analyses, respectively.

We split the dataset into training (70%) and test (30%) sets, ensuring each set has a balanced number of instances from both classes and trained Random Forest (RF) and Support Vector Machine (SVM) with radial kernel ML algorithms on the final feature set, with two times repeated 10-fold cross-validation and random search hyperparameter tuning. Table 3 shows the models' classification performances on the test dataset.

**Table 3.** Classification Performance of Models.

| Metric | **Accuracy** | Sensitivity | Specificity | AUC |
|---|---|---|---|---|
| RF* | **86%** | 100% | 71% | 94% |
| SVM** | **79%** | 100% | 57% | 96% |

*RF: Random Forest, **SVM: Support Vector Machine

## 4.2 Global Explanations

To assess the overall decision logic of models, we analyzed the dataset-level feature importance and PDP plots of the models.

The main idea of feature importance analysis is to measure how much a model's performance changes if each feature's values are permuted [85]. The result of this analysis on our dataset revealed that cross-correlation of HR-SpO2, sample asymmetry HR, and weight were the top 3 most important features for both models. Although separately, several previous studies had emphasized the importance of each of these features [51,56,66]. Therefore, the overall decision logic of the models may be considered plausible depending on this analysis.

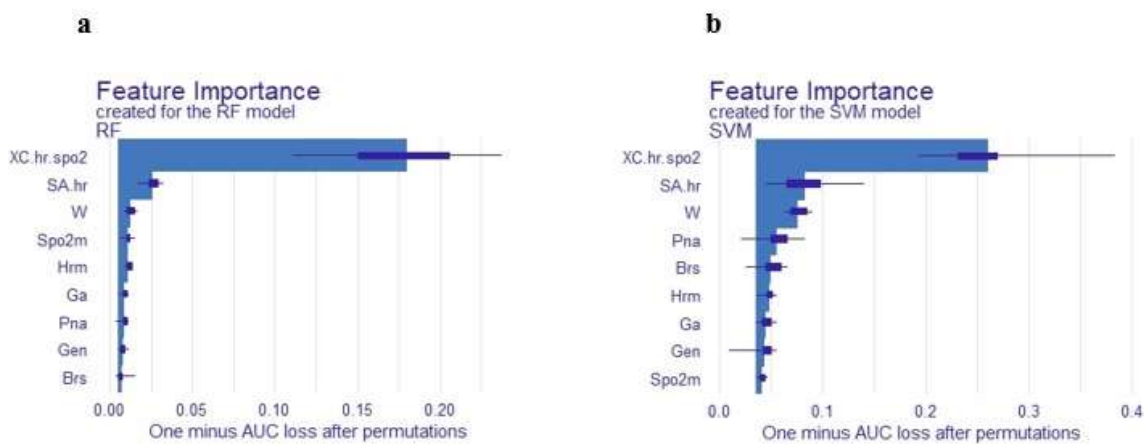

**Fig. 2.** Feature Importance analysis of RF (a) and SVM (b) Models.

We evaluated the one and two-dimensional PDPs for both models to assess how the model classification outcome behaves on average as a function of one or two features. For the one-dimensional PDPs, we considered only dynamic features of the entire set (Fig. 3). Although the models exploit features differently, the plots mostly revealed similar dependencies, particularly for the important features. As the cross-correlation of HR-SpO2 and the sample asymmetry of HR, the model prediction (from a medical perspective morbidity risk) also increases for both models. The risk increase trend is smoother in SVM PDP plots, while it is steeper in the RF PDP plots due decision boundary natures of the relevant algorithms. In the sample asymmetry of HR PDP for the RF model, due to the dense distribution of values around 0 (indicated by the marks on the x-axis), there is a slight increase in risk below 0 value. However, the classification outcome is still the same. The PDPs for the other dynamic features indicate that the trends for infants with either LOS are NEC. The increasing trend with a high mean Spo2 value is probably due to the oxygen support provided due to the treatment of the disease, but it is not causal.

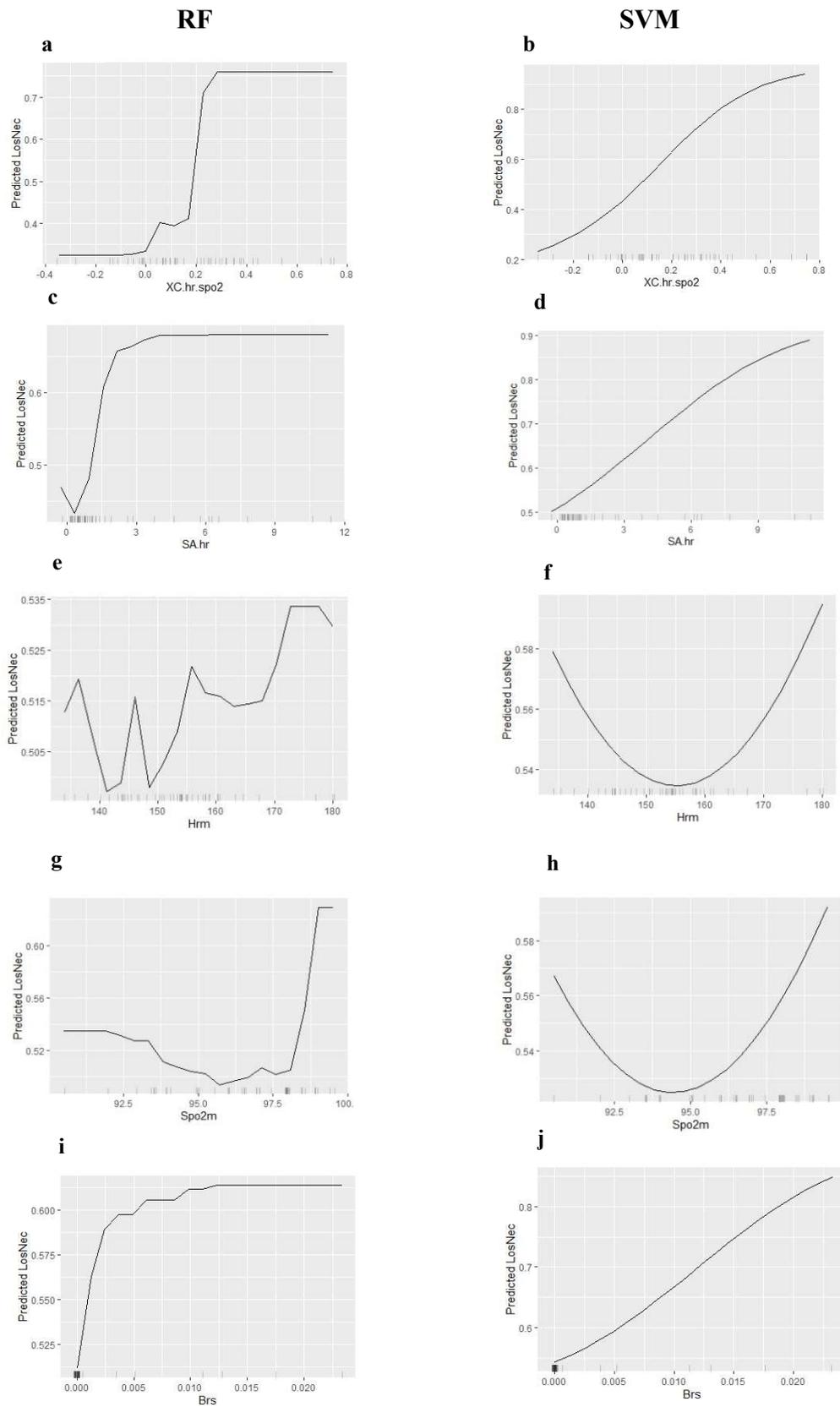

**Fig. 3.** Partial dependence plot (PDP) of dynamic features for RF (left) and SVM (right) models. Cross-correlation of HR-SpO2 (a and b), sample asymmetry of HR (c and d), mean HR (e and f), hypoxia percentage (g and h), and bradycardia percentage (i and j).

This fact contradicts the actual mechanism, as previously observed decrease of mean SpO2 in a study [56] and frequent desaturations are signs around the diagnosis. Although this feature is not an important variable for both models, this behavior must be considered in model improvement studies. The bradycardia percentage PDP also reveals an increase in risk with the increase of the feature values, with a steeper increase in a particular region for the RF model. There is no significant similarity in the mean HR PDPs of the models.

Fig. 4 shows the two-dimensional PDPs of weight against the cross-correlation of HR-SpO2 and sample asymmetry of HR features for the models. The color scale sidebar corresponds to the morbidity risk prediction of models. For the same dynamic feature values, the risk decreases as weight increases. Although the data distribution dominantly affects the color transition of the RF model, there is a smoother transition in SVM model PDPs. This picture is coherent with decreased HRV and immature ANS of preterm infants without a specific morbidity condition compared to healthy full-term newborns. This pattern also reminds the remark in [52] that calibrating neonatal sepsis decision support systems with weight measurements could increase predictive ability. It also indicates that the models capture the relations of static (weight) and important dynamic features to distinguish health and disease states. Assuring more data with a balanced distribution of gestational age and post-natal age [50] may improve this capability,

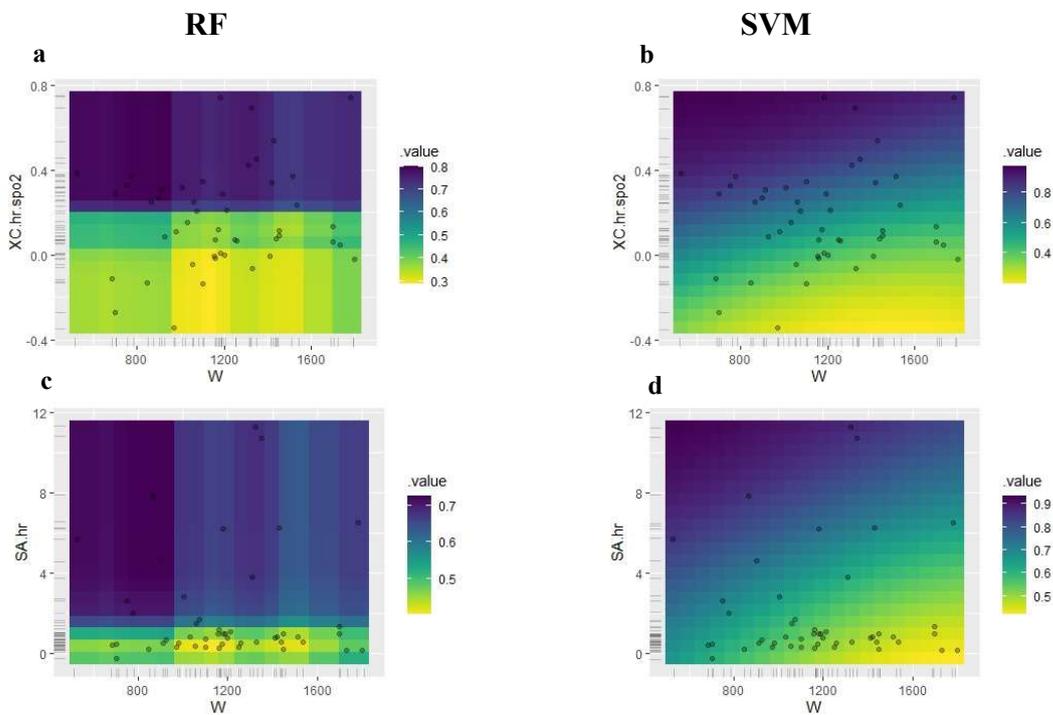

**Fig. 4.** Two-dimensional PDP of weight versus cross-correlation of HR-SpO2 (a and b) and sample asymmetry of HR (c and d) for RF (left) and SVM (right).

Nonetheless, due to the clinical coherence of one-dimensional PDP of important features and the observed prominence of weight along with dynamic features in two-dimensional PDP, we conclude that the overall decision logic of the models is reliable.

**4.3 Local Explanation and Misclassification Identification**

We randomly chose three cases from our dataset. Table 4 shows (dis)agreements regarding the models' classification decisions and the ground-truth classes of the cases.

**Table 4.** Model classifications and Ground-Truth classes of instances

|        | RF      | SVM    | Ground-Truth |
|--------|---------|--------|--------------|
| Case-1 | LosNec  | LosNec | Healthy      |
| Case-2 | LosNec  | LosNec | LosNec       |
| Case-3 | Healthy | LosNec | Healthy      |

We retrieved the ten most similar instances of latent space neighborhood for each case from our dataset using the Gower method. Due to the more dominant effects of immaturity on morbidity, we set the weight values as 2 for gestational age, weight, and post-natal age and 1 for gender. We prepared two-dimensional scatterplots, one of the dimensions being the similarity (Gower) distance from the case to the neighbors, and the other is one of the important dynamic features (Fig. 5).

The red points depict the case in question, and violets and greens depict the neighbors with and without either LOS or NEC, respectively. All the x-axes are Gower distances to the neighbors, and the y-axes are one of the important features (cross-correlation of HR-SpO2 or sample asymmetry of HR). We manually added horizontally gray lines to visualize the approximative feature value boundary between the two classes for this latent space neighborhood. Depending on the proximity of the Case's feature value to the boundary value, a clinician may reason the likelihood of the actual class the case belongs to.

When we apply this reasoning approach to our cases, we can infer that Case-1, Case-2, and Case-3 seem to belong to the 'Healthy', 'LosNec', and 'Healthy' classes, respectively, as stated in the Ground-Truth column of Table-4.

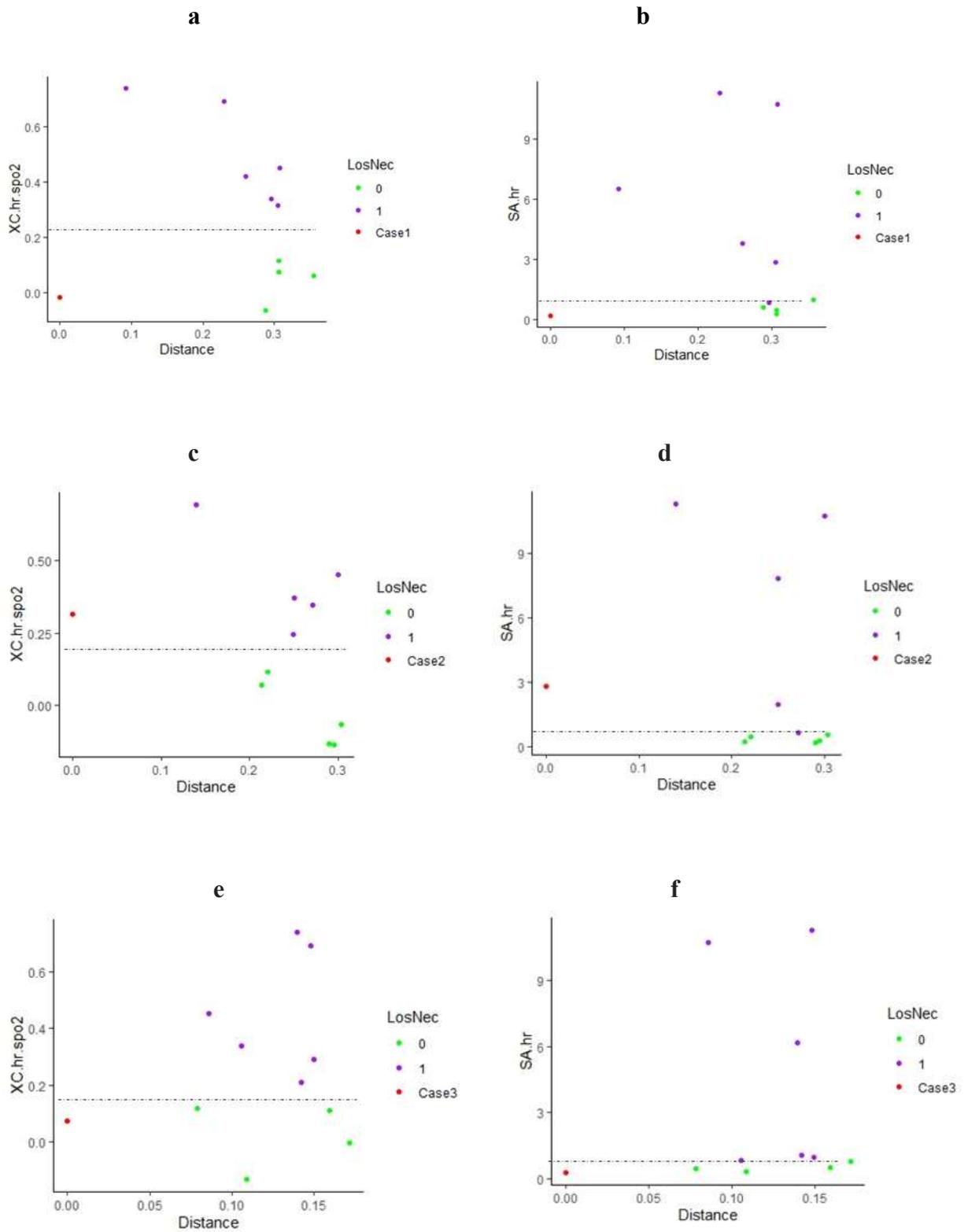

**Fig. 5.** LSS-based neighbors of Case-1 (a and b), Case-2 (c and d), and Case-3 (e and f). Horizontal lines depict the approximative feature value boundary between the two classes for this latent space neighborhood.

Considering our demographically balanced distributed dataset, these inferences are technical proof of concept for justification and contestation of the model decisions of the selected cases with a causal understanding. Besides, the visual interface aspect of our methodology is a facility, which has been stated as desired explainability approach for clinicians [72,74]. Thus, utilizing this facility, they make their final decision with autonomy by assessing similar patients' clinical details and trajectories.

## 4  Discussion

Recognition of algorithmic advice errors and improving the models are crucial aspects of trustworthy artificial intelligence in medicine [11–14]. Additionally, the importance of context and keeping the clinician in the loop have been highly recommended recently during the design of clinical decision support systems (CDSSs) [88–91]. Although there are several studies assessing algorithmic reliability considering these aspects in biomedical imaging domains [26,33–35], studies for tabular data are limited to global explanations and local model-agnostic methods [21] or case-based reasoning methods leveraging technical dimension reduction of the dataset [27] and important feature based similarity [28], with insufficient consideration of contextualization.

By fusing domain knowledge, the variational effect of maturity and gender on health and morbidity, we sought a novel idea of LSS-based explainability and misclassification identification of ML models trained with tabular medical data. We abstained from the algorithmic prediction based on latent space concepts, as the performance of such a model would depend on the regional data distribution of the dataset and may result in misleading conclusions. By quantifying and visualizing similarity within the latent space neighborhood, we focused on providing the clinicians with contextualized distilled information. Thus, they may decide with autonomy after assessing similar patients' clinical details and trajectories and either justify or contest the recommendation of an ML model.

The only XAI study of neonatal sepsis is for mortality prediction[63]. To our knowledge, this is the first explainability and contestability study of LOS and NEC morbidity recognition in the neonatology domain. We comprehensively evaluated the global and local explanation results from engineering and medical perspectives to investigate model failures and identify misclassifications. This is an essential good practice needed for trustworthy medical AI. We established a methodology with a causal approach and proof concept on our dataset to prevent risks of model failures of LOS and NEC recognition ML models for preterm infants, which was an open research question of the domain [62,70].

Considering the clinical context, our methodology can be translated to other medical domains as such latent patterns to distinguish health and disease states exist in other mechanisms of human physiology.

There are several limitations of our study. In most previous studies [46,56,59,61], the features were extracted from around-the-clock continuously monitored vital signs with sliding windows approach, while we used monitoring data of each one-hour interval for morning and afternoon when no care was given to infants. Nevertheless, the effect of sleep/awake states was not considered within the scope of those studies, although several previous research [92,93] state that HRV is affected by sleep state dynamics. Therefore, our dataset is free of the effect of these states. Our restricted sample size during model development may also be considered a limitation. Still, rigorous artifact cleaning, labeling, and overseeing the balance of demographic features ensures the quality of our dataset. In the future, we plan to increase the samples of our dataset, enrich the clinical features and evaluate other morbidities in the neonatology domain in addition LOS and NEC .

## 4. Conclusion

In this paper, we established an explainability and contestability methodology, which may be generally applied to medical AI models with consideration to clinical domain context. We demonstrated its effectiveness on a highly needed open research topic to prevent the risks of ML-based LOS and NEC recognition. Additionally, we presented the context-dependency, clinician autonomy, and model improvement concepts in medical AI's explainability and contestability framework. We also highlighted the importance of collaborative involvement of engineering and medical perspectives for trustworthy medical AI.

We hope our methodology will inspire the research community to apply, further investigate the effectiveness, and improve our approach in diverse medical domains using ML models trained with tabular data.